\documentclass[conference]{IEEEtran}
\IEEEoverridecommandlockouts
\usepackage{cite}
\usepackage[english]{babel}
\usepackage{amsmath,amssymb,amsfonts}
\usepackage{algorithmic}
\usepackage{graphicx}
\usepackage{textcomp}
\usepackage{xcolor}
\def\BibTeX{{\rm B\kern-.05em{\sc i\kern-.025em b}\kern-.08em
    T\kern-.1667em\lower.7ex\hbox{E}\kern-.125emX}}
    
\makeatletter

\def\ps@IEEEtitlepagestyle{
  \def\@oddfoot{\mycopyrightnotice}
  \def\@evenfoot{}
}
\def\mycopyrightnotice{
  {\footnotesize 978-1-6654-0009-1/21/\$31.00~\copyright~2021 IEEE\hfill} 

  \gdef\mycopyrightnotice{}
}

\@ifundefined{showcaptionsetup}{}{
 \PassOptionsToPackage{caption=false}{subfig}}
\usepackage{subfig}
\makeatother

\usepackage{eso-pic}
\newcommand\AtPageUpperMyright[1]{\AtPageUpperLeft{
 \put(\LenToUnit{0.5\paperwidth},\LenToUnit{-1cm}){
     \parbox{0.5\textwidth}{\raggedleft\fontsize{9}{11}\selectfont #1}}
 }}
\newcommand{\conf}[1]{
\AddToShipoutPictureBG*{
\AtPageUpperMyright{#1}
}
}

\title{Real Time Action Recognition from Video Footage}
\conf{2021 3rd International Conference on Sustainable Technologies for Industry 4.0 (STI), 18-19 December, Dhaka} 

\begin{document}

\author{
\IEEEauthorblockN{Tasnim Sakib Apon}
\IEEEauthorblockA{\textit{Computer Science and Engineering} \\
\textit{BRAC University}\\
Dhaka, Bangladesh \\
sakibapon7@gmail.com}
\and
\IEEEauthorblockN{Mushfiqul Islam Chowdhury} 
\IEEEauthorblockA{\textit{Computer Science and Engineering} \\ 
\textit{BRAC University}\\ 
Dhaka, Bangladesh \\ 
nnahid878@gmail.com} 
\and
\IEEEauthorblockN{MD Zubair Reza} 
\IEEEauthorblockA{\textit{Computer Science and Engineering} \\ 
\textit{BRAC University}\\ 
Dhaka, Bangladesh \\ 
zubairreza20@gmail.com} 
\and
\IEEEauthorblockN{Arpita Datta} 
\IEEEauthorblockA{\textit{Computer Science and Engineering} \\ 
\textit{BRAC University}\\ 
Dhaka, Bangladesh \\ 
arpita.datta.1998@gmail.com} 
\and
\IEEEauthorblockN{Syeda Tanjina Hasan} 
\IEEEauthorblockA{\textit{Computer Science and Engineering} \\ 
\textit{BRAC University}\\ 
Dhaka, Bangladesh \\ 
tzhasan00@gmail.com} 
\and
\IEEEauthorblockN{MD. Golam Rabiul Alam}
\IEEEauthorblockA{\textit{Dept. of Computer Science and Engineering} \\
\textit{BRAC University}\\
Dhaka, Bangladesh \\
rabiul.alam@bracu.ac.bd}

}

\maketitle

\begin{abstract}
Crime rate is increasing proportionally with the increasing rate of the population. The most prominent approach was to introduce Closed-Circuit Television (CCTV) camera-based surveillance to tackle the issue. Video surveillance cameras have added a new dimension to detect crime. Several research works on autonomous security camera surveillance are currently ongoing, where the fundamental goal is to discover violent activity from video feeds. From the technical viewpoint, this is a challenging problem because analyzing a set of frames, i.e., videos in temporal dimension to detect violence might need careful machine learning model training to reduce false results. This research focuses on this problem by integrating state-of-the-art Deep Learning methods to ensure a robust pipeline for autonomous surveillance for detecting violent activities, e.g., kicking, punching, and slapping. Initially, we designed a dataset of this specific interest, which contains 600 videos (200 for each action). Later, we have utilized existing pre-trained model architectures to extract features, and later used deep learning network for classification. Also, We have classified our models' accuracy, and confusion matrix on different pre-trained architectures like VGG16, InceptionV3, ResNet50, Xception and MobileNet V2 among which VGG16 and MobileNet V2 performed better.
\end{abstract}


\section{Introduction}
Recently, with the growth of the population, the crime rate is increasing day by day, and violent action is also occurring everyday. It can take place in your neighborhood, whereas reaching to school, at school, and at the office. There is no foolproof solution to control violent actions. Video surveillance provides a good role in real-time action recognition. Cameras are formed at each corner since the video surveillance system recognizes the scenes and detects divergent actions. Video crime detection is an essential subject in computer vision. In nearly every industry surveillance cameras are used. Consequently, due to the inefficiency of recordings, human monitoring on surveillance cameras is becoming redundant. Computer interference in management will substantially eliminate the problem of inactivity. It has become an important subject to make machines understand videos’ violent actions to imbrute the process. This paper introduces a novel technique on this subject and effectively enhances the quality of violent action video classification. The idea of violent action defines physical bullying which means harming any person by affecting his/her body. For violent action identification not much research has been completed. Our subsidies in this paper are, here, we will present the new datasets of violent action of fighting, slapping, punching, containing real-world fights. After that, we will propose embryonic procedures to intercept the fight detection problem in real-time, which will suffice as a touchstone for upcoming experimentation in the domain. \par
We had some limitations which we have faced regarding the completion of our project. (i) While doing our research, we faced hurdles in collecting our data as the required data according to the needs of our model-based were very difficult to find and after collecting,  the matching of data with our model was time lengthy and the process was difficult too. (ii) The field and topic which is the interest of our research, is very uncommon as we worked only with datasets of fighting scenes of kicking, punching, slapping, etc. which are abusive acts and identified them from video footages and we collected a good amount of data regarding our research purpose. There are very few works like papers and studies are very few regarding a similar kind of sector of our research area.
Contributions of our project are as follows:
\begin{itemize}
\item By constructing a pipeline of our dataset, we were able to acquire a large number of new dataset.
\item We attempted to construct our model in such a way that it would effectively operate with our dataset.
\item  We have used CNN, DNN instead of LSTM to find accurate and fruitful results for our model which will motivate future researchers who want to work regarding this study and will find more suitable ways to make it more perfect.
\end{itemize}
In our manuscript, section \ref{relatedwork} is all about the previous works which are related to our field of detecting from video footage, and section \ref{proposemodel} provides the information and description regarding our model which has three subsections. In section \ref{system}  we have described our model. In section \ref{data}, we talk about data acquisition and preparation. In section \ref{model}, we discuss about VGG-16, MobileNet V2, Xception, Inception V3, ResNet50 on our dataset and in Section \ref{performance}, after the comparison of results about the effectiveness of different models,  the better models with better accuracy have been found after the results of these accuracy results. And finally, in section \ref{conclusion}, the research has come to an end with the hope of better performance of our model and about the improvement in future works related to the same field.

\section{Related Work}  \label{relatedwork}

There are many evaluations that have been found which were used to detect fighting scenes or related topics using LSTM, KNN, or other pre-trained models whereas, we have used CNN \& DNN for our desired studies. In this section, we describe the previous work related to our field of research.\par

Despite having CCTV in several areas to protect people from any unexpected incidents, unfortunately, it is increasing rapidly because of its lack of effectiveness and they need a significant number of trained supervisors/operators \cite{perez2019detection}. In this study, the author examined some CCTV footage to determine how to improve the performance of explicit motion information. Here they proposed a pipeline through Two-stream CNN, 3D CNN, and a local interest point descriptor. To start this project, they collected 1000 videos of real-world CCTV-Fights and observed them by their characteristics. Additionally, they used the Visual Feature Extraction as the first stage, which is structured with RGB information. Then the author used a two-stream-based approach (RGB and Optical Flow) performed by 2D-CNN architecture, a 3D-CNN pipeline (with temporal information as the third dimension) using convolutional neural network architecture, and local interest points. They chose to mean average precision (mAP, higher is better) to determine the performance of these given methods. After applying the methods in the CCTV-Fights dataset, the percentage of mAP of Two-Stream is higher than any other method, an indicator of performance improvement. Moreover, they examined CCTV and Non-CCTV video and they observed that the Non-CCTV fights have a higher mAP than CCTV fights. From these experiments, they realized that explicit motion information is more auspicious than the RGB-only methods. \par

     Due to the lack of automatic detection of violent scenes, the children can watch violent movies and scenarios on the internet which is a matter of concern for a parent \cite{gong2008detecting}. In this study, the author tried to automatically detect violent videos to prevent the children from watching them. Here, the author selected three stages: segmentation of video into a set of shots, training in a semi-supervised way of low-level visual and auditory features, identifying the high-level audio effects, and determining whether the video is violent using probabilistic output stages. Moreover, the author had used two different algorithms like SCFL and SVM. Here SCFL is a semi-supervised cross-feature learning algorithm utilized in learning the classifier. On the other hand, SVMs are trained to classify audio clips to specify the audio effect each to each SVM model. To evaluate the difference between the SCFL and SVM model the author has experimented with the datasets over several movies and using precision, recall, F1-measure the author identified that the performance of SCFL is better than SVM by a significant margin. Finally, the author claimed that using SCFL the violence of a video can be measured automatically. \par
     In today’s generation, the spreading of violent videos throughout the internet is very common which should have to be prevented automatically without human involvement  \cite{foinivideo}. In this paper, the authors initially observed that CNN is a better approach, which extracts the fine-grained visual details via filters from the static images, which results in an excellent performance in 2D image classifications. The same performance is achieved for videos by using 3 dimensional CNN filters. However, for both cases, the issue arrives as the size of the dataset. Because a substantial amount of training and diverse samples are needed for training deep neural nets. To tackle this problem, the authors proposed a new dataset named Foi-Fight dataset with non-violent content and also gathered the violent video datasets from different sources like VISILAB (600 videos), UCF crimes (128-hour videos), CCTV-Fight, etc. Moreover, he invented a new implementation strategy named zoom to increase the neural network’s capacity so that it can speed up the handling of the number of frames and make the classification more careful and precise. To use this strategy, he used 3 inputs i) RGB input for single-stream I3D, ii) RGB and RGB difference input for two-stream I3D and iii) optical flow input for two-stream I3D through LiteFlowNet. After training those datasets with his new strategy he checked the accuracy using T-Accuracy, FN-Accuracy, and FP-Accuracy. Finally, he determined that zoom (his invented strategy) is better than the previous strategies to automatically inspect violent videos from movies or web videos. \par
     
    In the follwoing paper, the author focused on finding higher accuracy in fight detection\cite{nievas2011violence}. They came up with a new dataset of hockey videos to find the violence in sports footage. Here the author used the bag-of-words approach with using the datasets of hockey game videos of National Hockey League, INRIA (contains kicking or punching video), CAVIAR (instance people aggressive behavior videos) to detect aggressive violence. They used two prominent Spatio-temporal descriptors to examine video violence named STIP and MoSIFT using HOG, HOF, HNF features vectors. After the performance using HOG, HOF, and HNF vectors they found that MoSIFT has higher performance than STIP. Finally, they observed that detecting violence in hockey footage is easier than detecting fights in movies or other actions, and using the bag-of-words approach in MoSIFT can get approximately 90\% accuracy. \par  
Table \ref{relatedworkcomparison} shows the comparison between the findings of different conducted studies and our studies. 

\begin{table}[!t]
\caption{Comparison between different studies.}
\begin{center}
 \begin{tabular}{|c |c |c |c| |c} 
 \hline
 Model & Research Area & Accuracy & Reference \\ [0.5ex] 
 \hline\hline
CNN + LSTM & CCTV-Fight  & 79.5\% & \cite{perez2019detection}\\ 
 \hline
 I3D Network(RGB) & Foi-Fight &90.97\% & \cite{foinivideo}\\
 \hline
 FightNEts & Mini Foi-Fight & 70.94\% & \cite{foinivideo}\\
 \hline
  {BoW(MoSIFT)} & {Action Movies} & {80\%} & {\cite{nievas2011violence}}\\
 \hline
 SCFL & Action Movies & 88.71\% & \cite{gong2008detecting}\\
  \hline
  Motion & Hockey game & 84.5\% & \cite{fu2015automatic}\\
  Analysis Algorithm & & & \\
  \hline

 VGG-16 & ImageNet & 87.15\% & \cite{suspicious}\\
  \hline
 Differenct Pre-  & Manually Collected & 92.3\% & Present In \\
  trained model & Data & & This Study\\
 \hline
\end{tabular}
 \label{relatedworkcomparison}
\end{center}
\end{table}

\section{Proposed Real Time Action Recognition Method}  \label{proposemodel}
From Section \ref{proposemodel}, we get clear information about our proposed model and its efficacy which is divided into three parts. In part \ref{system}, we describe our system model, and then in part \ref{data}, we mention about our data acquisition, framerate and array conversion to images before feeding them into the Deep Neural Network model and finally in part \ref{model}, we discuss about the pre-trained models which we have used for our desired model.

\subsection{System Model} \label{system}
\begin{figure*}[!t]
\centering
\includegraphics[scale=0.52]{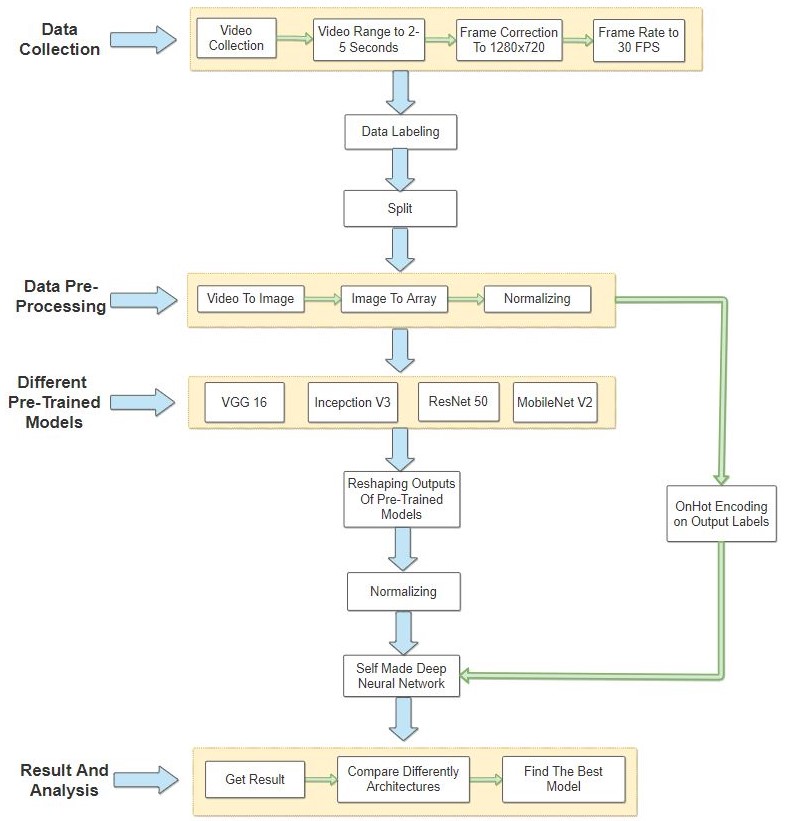}
\caption{Proposed System Model.}
\label{fig:x workflow}
\end{figure*}

We primarily focused on collecting data for three distinct categories, slapping, punching, and kicking. These data were collected mainly from the public video-sharing domain, i.e., YouTube, and sampled manually (the frame regions only corresponding to the actions of our interest). We ensured high-definition resolution which was 1280 x 720 of the videos and a unified frame rate of 30 for all clips. On average, each clip duration ranged from 2 to 7 seconds, which indicates the only period for the actual action. Further, we have labeled all the videos manually which was a little bit slow process but very efficient in assigning the right labels. After splitting the videos into train, test, and validation, we have extracted a few frames from each video. Then we have converted those frames into arrays and normalize the values we get from the images. At the same time, we used one-hot encoding on our labels. After finishing this pre-processing part we feed these values into some pre-trained models and get an array of values for each image. After that, we again normalized those values to get better solutions. To feed these values into our model we reshape them into 1-dimensional shapes. On our model, we tried to generate outputs the same as our one-hot encoded label outputs. The next step was to compare which pre-trained model is generating good answers for our model. In short, we tried to create a solution where we can use existing solutions along with our solutions to get better results. Our Deep Neural Network model contains 5 fully connected layers. Input layer of this model takes normalized output value of pre-trained model. We have used dropout  with 4 layers. Dropout is a method
of dealing with overfitting. The main concept of dropout is dropping a unit randomly during training from the neural network. We set the rate of dropout to 0.5, which ensures that in each epoch, 50\% of the neurons in that layer would be lowered at random. If there were 512 units in the fully connected layer, then only 256 would be trained in the second fully connected layer after 50\% of the neurons are removed. The 256 neurons are chosen at random and dropped. In our self design Deep Neural Network model we used two activation functions. One is ReLU and another one is softmax. We use the softmax activation function in output layer. We apply the softmax activation function in our model to convert real value into probability. Then we compared probability values for few frames and predicted the action. The softmax activation function assists us in obtaining more accurate results. Later for testing we first convert the videos into frames and than we loop through each frame and store their prediction into a list. Finally based on the majority we take our decision.

\subsection{Data Acquisition and Preparation}  \label{data}
Video is a collection of images. By the term video we understand recording, reproducing, or broadcasting of moving visual images. Like images, videos also have
width, height and depth. As video is a representation of visual images, we need to decide at which rate images will be shown in a video. We mention its rate as frame rate per second, in short fps. Fps refers to how many images will be shown in a video per second. Working with all frames
of a video might not be a good solution for our problem as we are thinking of predicting a real-time situation where working with each frame might lead to a wrong assumption and this process is more time-consuming. So, we have divided the frame rate with each frame number and taken only those frames which set the reminder value to zero. Entire process has been done using openCV3. Initially, our frame size was (1280×720). After doing some research we found out that most of the pre-trained model takes (224,224,3) shape as input. While loading frames, we converted our image into this shape. After converting to an array we normalized all the image arrays to extract more
information from it. After extracting feature we have gain normalized our input shape before feeding into the Deep Neural Network model.
As previously mentioned, in our work we have detected 3 types of actions. Which are kick, slap and punch. At first, we manually assigned labels for each video data. While converting video to image we stored the label with the frame by including action to the frame name. We have used one hot encoding to create output classifications. We have created a separate output row for each label and populated it with binary 0 and 1 according to the previously created labels.

\subsection{Model Specification} \label{model}

\subsubsection*{Inception V3}
\begin{figure}[!t]
\centering
\includegraphics[scale=0.33]{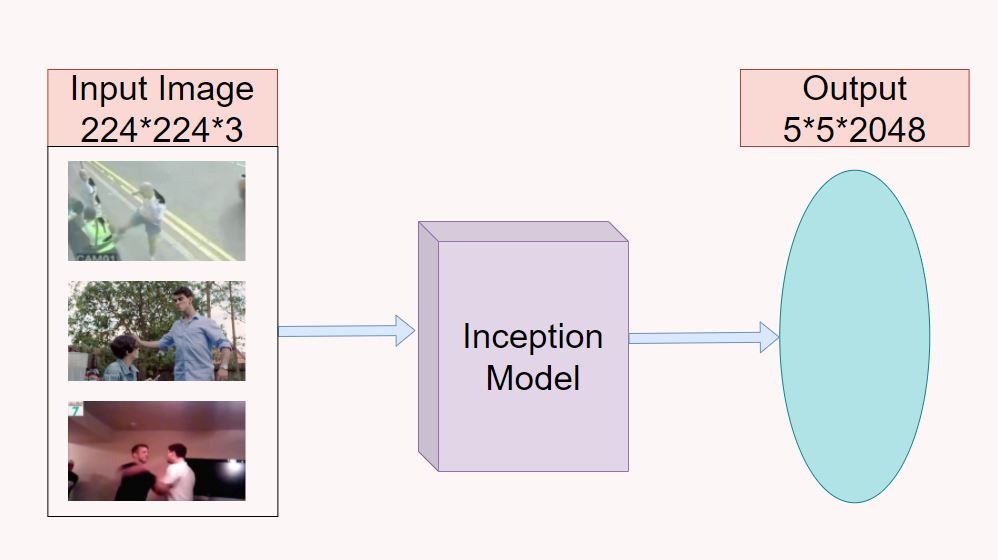}
\caption{Basic Model Architecture with Inception V3.}
\label{fig:x inceptionarchi}
\end{figure}

The inception model is a very simple yet powerful architectural unit. The key idea behind this model is the inception block. The purpose of this model is to act as a multi-level feature extractor. It means it will be able to compute 5×5, 3×3 and 1×1 convolutions within the same module. Before being fed into the next layer the outputs of these filters are stacked along the channel dimension\cite{aljuhani1989going}. Firstly, they factorized convolutional which helps to reduce the computational efficiency. Then build smaller convolutions that replace bigger convolutions. Here they focused on unbalanced convolution where a 3 × 3 convolution could be replaced by a 1 × 3 convolutional followed by a 3 × 1 convolution. Then focused on auxiliary classifiers which are small CNN inserted between layers during training. Finally, they reduced the grid. size\cite{szegedy2016rethinking}. Fig \ref{fig:x inceptionarchi} shows the architecture of Inception V3

\subsubsection*{ResNet50}
Deep Convolutional neural networks are generally good at identifying features from images. Also stacking more layers provides higher accuracy. So the idea is that shouldn’t building better neural networks as easy as adding more layers to the network. However, the authors of resnet50 states that if you just continue to concatenate convolutional layers on top of activations and batch normalization the training will eventually get worse, not better \cite{he2016deep}. To address this problem the authors came up with a deep residual learning framework. ResNets are built out of residual blocks. If you consider architecture and its deeper counterpart with more layers, theoretically all the deeper models will copy the output from the shower model with identity mapping. So the suggested solution is that the deeper model should not produce higher error than the shallow counterpart. So, therefore the residual functions formulate the layers having a connection with the input through identity connections. Now, it can easily push the layer down to zero. This is eight times deeper than VGG nets but in terms of the floating-point operation measurement, it actually has less computation.  Here our input shape was the same as before and output shape was 7×7×2048. 

\subsubsection*{VGG-16}
\begin{figure*}[!t]
\centering
\includegraphics[scale=0.33]{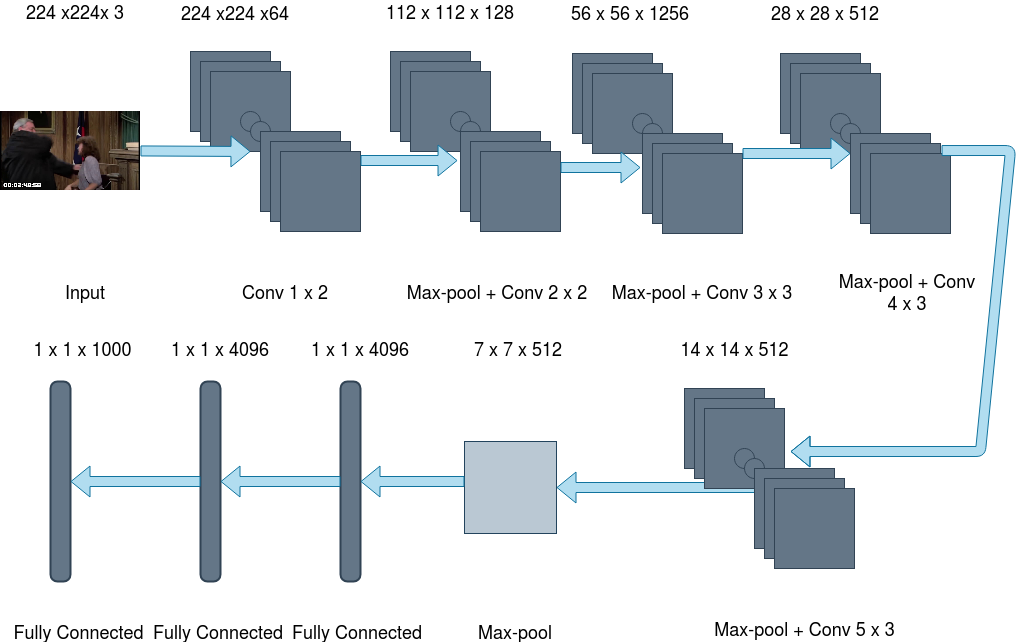}
\caption{Model Architecture with VGG-16. }
\label{fig:x vggarchi}
\end{figure*}

The (VGG) Visual geometry group a well-known DCNN show, which was excerpted by K.Simonyan and A. Zisserman in 2014\cite{simonyan2014very}. The essential VGG-16 structure is shown in Fig \ref{fig:x vggarchi}.

VGG formed 92.7\% as the top 5 test precision at OLSVRC competition. The most important part is expanding the profundity of the organize with exceptionally small 3 × 3 convolution channels by two convolutional layers are utilized ceaselessly with a corrected straight unit ReLU as actuation work taken after by a max-pooling layer, a number of completely associated layers with ReLU and soft-max as the ultimate layer. VGG Net has three categories depending on the total number of layers existing within the engineering, they are VGG-11, VGG16 and VGG-19. VGG-16 and VGG-19 can be 16 and 19 layer projects respectively. This meaning The VGG-16 design consists of 16 convolutional layers and the VGG-19 consists of 19 convolutional layers. Compared to the "VGG-19" organization program, the "VGG-16" organization program has less weight. The "VGG-16" metrics and partial counts are closely related to the classifier and discard layer regularization. We ha used output of max-pool which is 7×7×512.

\subsubsection*{MobileNet V2}
\begin{figure*}[!t]
\centering
\includegraphics[scale=0.45]{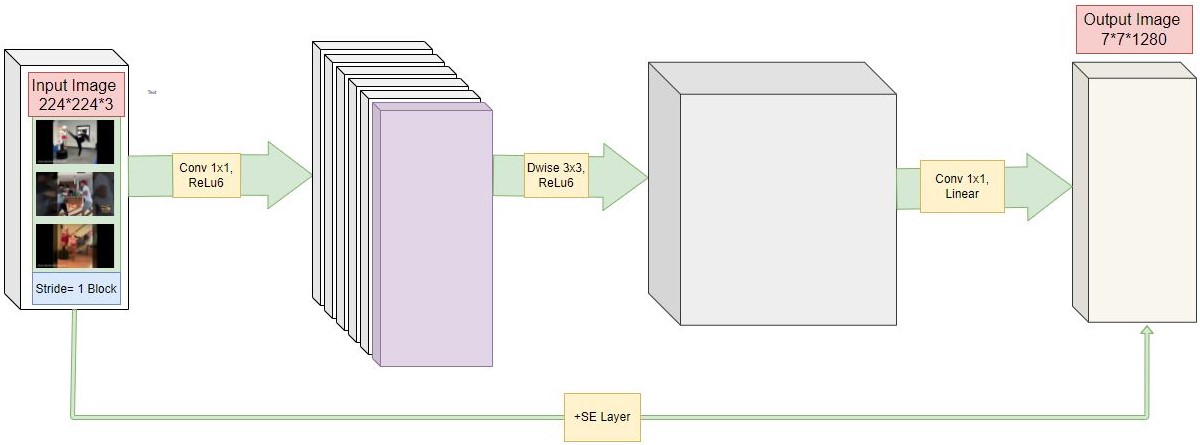}
\caption{Basic Model Architecture With MobileNet V2.}
\label{fig:x mobilenetarchi}
\end{figure*}

MobileNet is a class of lightweight deep convolutional neural networks \cite{howard2017mobilenets}. It uses depth-wise separable convolutions. It is about 10x faster than VGG-16 and 3x faster than the Inception image classification pre-trained model. Despite being fast it is very small in size. MobileNetV2 was introduced in 2019 by google researchers which was an improved version of MobileNet. The authors added new layers in its main building block. The authors added expand layers, projection layers, and residual connections to its architecture. This time they used three convolutional layers. The first layer is the 1×1 convolution layer also known as the expansion layer. The last two convolutional layers are depthwise convolution, which filters the inputs, and the 1×1 pointwise convolution layer. However, the 1×1 layer makes the number of channels smaller also known as the projection layer \cite{sandler2018mobilenetv2}. Fig \ref{fig:x mobilenetarchi} shows MobileNetV-2 architecture.

\subsubsection*{Xception}
Xception model was built on the inception model which was introduced by google researchers. It is a deep convolutional neural network with depth wise separable convolutions. It has performed better than most of the popular pre-trained models like ResNet50, VGG-16, Inception. This model is divided into three parts. At first the data is passed through the enty flow which is the first part. Then it passes through the middle part which is repeated eight times and then finally it enters the exit flow \cite{chollet2017xception}. We used the basic xception model which was available in keras. We fed our image with a shape of 224×244×3 and received output shape of 7×7×2048.
\begin{figure}[!t]
\centering
\includegraphics[scale=0.42]{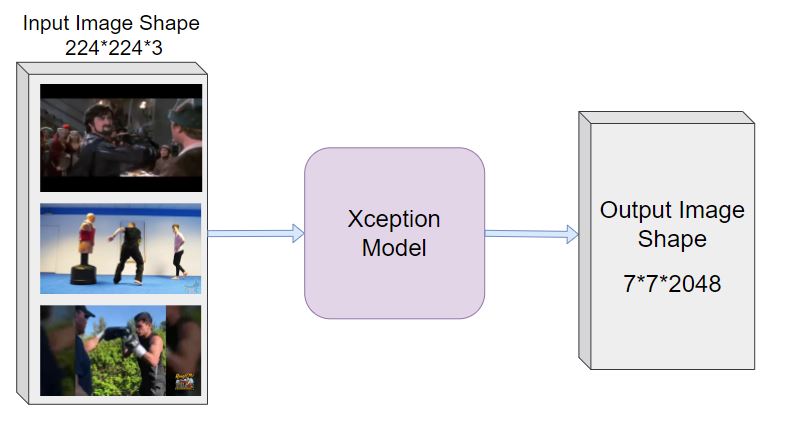}
\caption{Basic Model Architecture With Xception.}
\label{fig:x xception}
\end{figure}

\section{Performance Evaluation} \label{performance}

\begin{table}[!t]
\caption{Model Performance I.}
\begin{center}
 \begin{tabular}{||c |c |c |c| |c|} 
 \hline
 Model & Precision & F-1 Score & Recall \\ [0.5ex] 
 \hline\hline
 VGG-16 & 89.6 & 89.6 & 89.6\\ 
 \hline
 Inception V-3 & 88.6 & 88.3 & 88.3\\
 \hline
 ResNet50 & 69 & 68 & 68\\
 \hline
 \textbf{MobileNet V-2} & \textbf{92.3} & \textbf{92} & \textbf{92.3}\\
 \hline
 Xception & 91.6 & 91.6 & 91.6\\
 \hline
 \hline
\end{tabular}
 \label{modescorefinal}
\end{center}
\end{table}

\begin{table}[!t]
\begin{center}
\caption{Model Performance II.}
\begin{tabular}{ ||c||c||c|c|c|| } 
\hline
Model & Action & Precision & F-1 Score & Recall \\
\hline \hline
 & Kick & 0.92  & 0.93 & 0.92 \\
VGG-16 & Punch & 0.89  & 0.87 & 0.90\\ 
& Slap & 0.88  & 0.89 & 0.87\\ 
\hline
 & Kick & 0.91  & 0.92 & 0.92 \\ 
Inception V-3 & Punch & 0.88  & 0.87  & 0.88\\ 
& Slap & 0.87  & 0.86 & 0.85\\ 
\hline
 & Kick & 0.73  & 0.70 & 0.67\\ 
ResNet50 & Punch & 0.74  & 0.74 & 0.60\\ 
& Slap & 0.60  & 0.60 & 0.77\\ 
\hline
 & Kick & 0.93  & 0.92 & 0.93 \\ 
\textbf{ MobileNet V-2} & Punch & 0.93  & 0.91 & 0.93\\ 
& Slap & 0.91  & 0.93 & 0.91\\ 
\hline
 & Kick & 0.94  & 0.93 & 0.94 \\ 
 Xception & Punch & 0.92  & 0.92 & 0.92\\ 
& Slap & 0.89  & 0.90 & 0.89\\ 
\hline \hline
\end{tabular}
 \label{modescore}
\end{center}
\end{table}

For testing purposes we first made sure that our data is in 30fps. If not, we first convert our data and than convert the video into frames and pick 1 frame in every second. The purpose of converting our data into 30fps is to ensure that we only get 1 frame in each second. Than we loop through each frame and store their prediction into a list. Finally based on the most prediction we pick our decision. Through this technique we have tested with five different image classification pre-trained models. Table \ref{modescorefinal} depicts each pre-trained model's final score. The best score was 92.2\% for MobileNet V-2 . With 91.6\% Xception, 89.6\% with  VGG-16 and 88.4\% with Inception V-3 also performed better than rest of the models. Resnet50 however had the lowest score which is 68.3\%. The reason behind ResNet50's accuracy fall is because of its deep architecture. Each of the pre-trained model's precison, F-1 score and recall sore for each class is presented in Table \ref{modescore}. Here the best fitted model, MobileNet V2 do not have any fluctuation in its precision, recall value for any of its available classes thus F1-score is stable. Except ResNet50 the other pre-trained models were too stable and ResNet50 however have low f1-score for detecting slap. Previously with the similiar kind of dataset Foi-Fight, I3D Network(RGB) model was able to achieve an accuracy of 90.97\%. Our Model outperformed the previous model's accuracy with our proposed dataset. After comparing all the models we can state that MobileNet V-2, Xception, VGG-16 and Inception V-3 were more stable models with better accuracy.

\section{Conclusion} \label{conclusion}
Our experiment model, which we have presented, will detect violent action from the video footage from a large dataset to find out the violent activities that happened with victims. Though the background noise is a big challenge, it will be able to perform better with high maintenance and it will help detect the violent  activity, which will help the security organizations take actions against violent actions, which can turn into big crime. Our model can detect violent actions with the existing dataset, and we will make more progress to our model. In the future, we plan on generating more data to create a larger dataset, and we also plan on using another machine learning algorithms such as LSTM, decision tree classifiers, AdaBoost classifiers, K neighbors classifiers, random forest classifiers, etc to improve our model to work more efficiently.

\vspace{12pt}

\end{document}